%% file: main.tex
\def\year{2021}\relax
\documentclass[letterpaper]{article} 
\usepackage{aaai21}  
\usepackage{times}  
\usepackage{helvet} 
\usepackage{courier}  
\usepackage[hyphens]{url}  
\usepackage{graphicx} 
\def\UrlFont{\rm}  
\usepackage{natbib}  
\usepackage{caption} 
\usepackage{pifont}
\usepackage{amsmath}
\usepackage{tabularx}
\usepackage{booktabs}
\usepackage[flushleft]{threeparttable}
\usepackage{comment}
\usepackage{xcolor}
\usepackage{amsfonts}
\usepackage{multirow}
\usepackage{fontawesome}

\newcommand\papername{\textsc{PARA-COMET}}
\newcommand\model{\textsc{PARA-COMET}}

\newcommand{\cmark}{\ding{51}}%
\newcommand{\xmark}{\ding{55}}%

\newcommand{\comet}{\textsc{COMET}}

\definecolor{darkspringgreen}{rgb}{0.09, 0.45, 0.27}

\newcommand{\correct}{\textcolor{darkspringgreen}{\cmark}}
\newcommand{\incorrect}{\textcolor{red}{\xmark}}

\newcommand\ai{$^\diamondsuit$}
\newcommand\uw{$^\spadesuit$}

\title{Paragraph-level Commonsense Transformers with Recurrent Memory}
\author{Saadia Gabriel\uw\ai\space\space\space Chandra Bhagavatula\ai\space\space\space  Vered Shwartz\uw\ai\space\space\space Ronan Le Bras\ai \\ Maxwell Forbes\uw\ai\space\space\space Yejin Choi\uw\ai \\
}
\affiliations{
\uw Paul G. Allen School of Computer Science \& Engineering, University of Washington, Seattle, USA \\ 
\ai Allen Institute for Artificial Intelligence, Seattle, USA\\
\{skgabrie, mbforbes, yejin\}@cs.washington.edu , \{chandrab, vereds, ronanlb\}@allenai.org\\

}

\begin{document}

\maketitle
\begin{abstract}
\input{00-abstract.tex}

\end{abstract}


\begin{figure}[!t]
    \centering
    \includegraphics[width=1\linewidth]{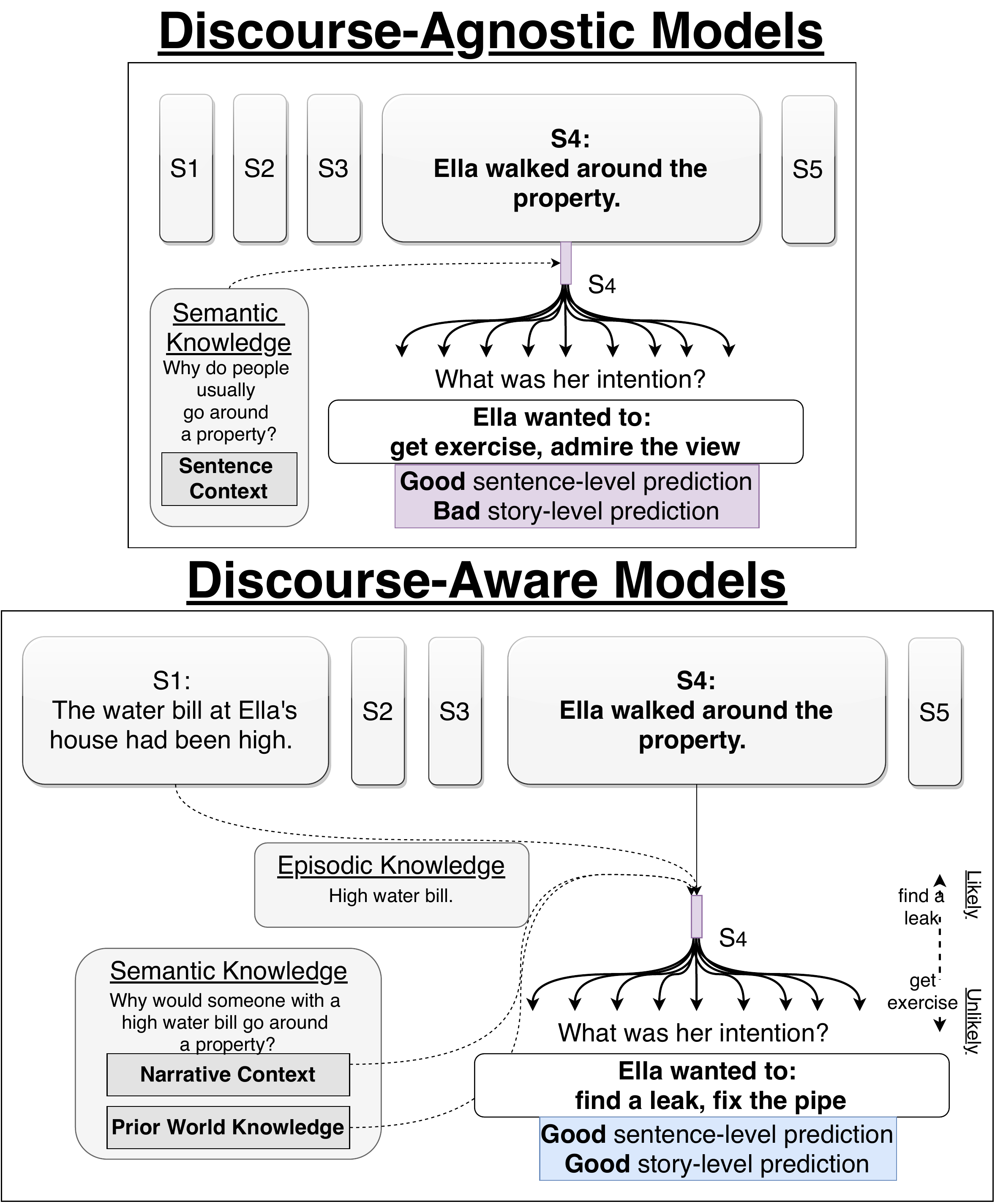}
    \caption{Discourse-agnostic models generate inferences relevant to the local context, but these generations can often be generic or incorrect at the narrative-level. Discourse-aware models take the rest of the context into account to make globally coherent inferences.  
    }
    \label{fig:nkg}
\end{figure}

\section{Introduction}
\label{sec:intro}
\input{01-intro.tex}

\section{Background}
\label{sec:background}
\input{02-background.tex}

\section{Discourse-Aware Commonsense Inference}
\label{sec:task}
\input{03-task.tex}

\section{Distant Supervision Approach}
\label{sec:data}
\input{04-data.tex}

\section{Model} 
\label{sec:model}
\input{05-model.tex}

\input{figures/human_eval_results}

\section{Experimental Setup}
\label{sec:experimental_setup}
\input{06-experimental_setup.tex}

\section{Evaluation}
\label{sec:experiments}
\input{07-experiments.tex}

\section{Case Study: Personal Narratives}
\label{sect:case_study}
\input{08-case_study}


\section{Conclusion}
\label{sect:conclusion}
\input{10-conclusion}

\section{Implications and Ethics Statement} 
\label{sect:ethics}
\input{11-ethics}

\section*{Acknowledgments}
We thank the anonymous reviewers for helpful feedback, as well as Maarten Sap, Hannah Rashkin, Eunsol Choi and members of the UW and AI2 communities for helpful discussions. This research was supported in part by DARPA under the CwC program through the ARO (W911NF-15-1-0543) and DARPA under the MCS program through NIWC Pacific (N66001-19-2-4031).

\bibliography{anthology,references}

\clearpage
\appendix
\input{12-appendix}

\end{document}

%% file: 00-abstract.tex
Human understanding of narrative texts requires making commonsense inferences beyond what is stated explicitly in the text. A recent model, \comet{}, can generate such implicit commonsense inferences along several dimensions such as pre- and post-conditions, motivations, and mental states of the participants. However, \comet{}~was trained on commonsense inferences of short phrases, and is therefore discourse-agnostic. When presented with each sentence of a multi-sentence narrative, it might generate inferences that are inconsistent with the rest of the narrative. 

We present the task of discourse-aware commonsense inference. Given a sentence within a narrative, the goal is to generate commonsense inferences along predefined dimensions, while maintaining coherence with the rest of the narrative. Such large-scale paragraph-level annotation is hard to get and costly, so we use available sentence-level annotations to efficiently and automatically construct a distantly supervised corpus. 

Using this corpus, we train \model{}, a \emph{discourse-aware} model that incorporates paragraph-level information to generate coherent commonsense inferences from narratives. \model{} captures both \emph{semantic} knowledge pertaining to prior world knowledge, and \emph{episodic} knowledge involving how current events relate to prior and future events in a narrative. Our results show that \model{} outperforms the sentence-level baselines, particularly in generating inferences that are both coherent and novel.

%% file: 01-intro.tex
Narrative understanding is a long-standing challenge in the field of natural language processing (NLP) \cite{charniak1972toward,winograd1972understanding}. 
Arguably, the most crucial aspect of narrative understanding is the ability to make implicit commonsense inferences about entities and events in a story and refining them as the story unfolds \cite{Narratives,Williams2017UnderstandingSW,Rashkin2018ModelingNP,Qin2019CounterfactualSR}. This ability in humans is seamless, yet essential for coherent understanding of narrative text. \textit{Can NLP systems explicitly generate commonsense inferences, that a human might implicitly make while reading a narrative?}

Being able to generate commonsense inferences has important practical implications. Commonsense Transformer \citep[\comet,][]{Bosselut2019COMETCT}, proposed recently, generates  commonsense inferences for a given phrase or sentence, capturing pre- and post-conditions along nine inferential dimensions found in the ATOMIC \cite{Sap2019ATOMICAA} knowledge base.\footnote{See Table \ref{table:atomic_dimensions} for a full list of inferential dimensions in ATOMIC.} 
The commonsense inferences generated by \comet{} have been effectively applied to downstream applications such as sarcastic comment generation~\cite{chakrabarty-etal-2020-r}, therapy chatbots~\cite{Kearns2020AWI}, abductive natural language generation~\cite{bhagavatula2019abductive}, and automated story plot generation~\cite{ammanabrolu2020automated}.

However, the \comet{} inferences suffer from a major shortcoming -- they are generated for a sentence in isolation and fail to account for the full \textit{paragraph-level} narrative context. This often results in the generation of inferences that are inconsistent or unlikely when considering the previous narrative context. For example in Figure~\ref{fig:nkg}, given only the sentence \textit{``Ella walked around the property,''} one might infer that she did this because she wanted to \textit{``get exercise''} or \textit{``admire the view''}. While such an inference is reasonable for the sentence in isolation, it is inconsistent given the full context -- e.g., \textit{``The water bill at Ella's house had been high. Ella walked around the property.''} Instead, a more reasonable inference in light of the full context is that \textit{``She wanted to fix a leak.''}


We introduce the task of generating implicit discourse-aware commonsense inferences for narrative text, and present \model{}, a transformer-based, controlled generation model for the task. 
Instead of collecting crowdsourced annotated data as direct supervision for this task, which is potentially expensive and challenging to scale, \model{} is distantly supervised through sentence-level inferences obtained either from the \comet{} model or by heuristically matching a sentence to events found in the ATOMIC knowledge base. We define and use a coherence metric that measures the likelihood of each candidate inference in the context of the story to improve their paragraph-level consistency. 

We show that \papername{} generates coherent discourse-aware inferences and performs better than discourse-agnostic baselines in both automated and manual evaluation.
Yet, even the best model generates implausible inferences (23\% of the inferences), and inferences that contradict the paragraph-level context (in 44\% of the stories). This stresses the difficulty of the task and calls for further research. We release our models and data as an initial step towards advancing paragraph-level commonsense understanding.\footnote{Code and data is available at \url{https://github.com/skgabriel/paracomet}.}

%% file: 02-background.tex
\input{figures/task_examples}
\input{figures/atomic_dimensions}

\paragraph{Sentence-level Commonsense Inferences.} A key component of our distant supervision approach is the availability of sentence-level commonsense inferences. The ATOMIC knowledge base \cite{Sap2019ATOMICAA} consists of such \textit{if-then} knowledge about causes and effects, agents and themes of events, and their actions and mental states. An ATOMIC entry is encoded as a triplet $<e_1, d, e_2>$, where $e_1$ is an event phrase, $d$ is an inferential dimension and $e_2$ is the inference along the given dimension. ATOMIC defines nine inferential dimensions such as \texttt{xIntent}: the agent's intent, \texttt{oEffect}: the effect on the patient(s) etc. (See Table~\ref{table:atomic_dimensions}). The event $e_1$ and the inference $e_2$ are natural language templates consisting of variables \texttt{PersonX} for the agent and \texttt{PersonY} for the (possibly unknown) patient(s).\footnote{We refer to \texttt{PersonY} in ATOMIC as \textit{patient}, one or more people who are affected or acted upon by the action of the verb. We don't make the semantic distinction between patient and theme.} 

While ATOMIC contains nearly 880K triplets, it is not nearly enough to capture the full range and generality of possible events, which is immeasurably vast and impossible to manually enumerate. Furthermore, due to lexical variability, events are rarely found as-is in ATOMIC. To that end, \comet{} \cite{Bosselut2019COMETCT} was developed as a transformer-based knowledge model trained on ATOMIC to generate commonsense inferences for a given phrase/sentence. Thus, both ATOMIC and \comet{} are natural candidates to obtain \textbf{\textit{sentence-level}} commonsense inferences.

\paragraph{Reasoning about narratives.} A related line of work to ours is \emph{script learning}, that defines a structured representation for prototypical series of events \cite{schank1977scripts}. An event (e.g., going to a restaurant) is decomposed into components such as the participants (customer, waiter, cook, etc.), subevents (sitting down, asking for menus, etc.), and their various pre- and post-conditions. In later work, scripts were also referred to as ``narrative event chains'', and multiple methods to learn the narrative chains from raw text were developed \cite{chambers2008unsupervised,jans2012skip,pichotta2014statistical,rudinger2015script}. Similarly, the \emph{Choice of Plausible Alternatives} (COPA) task \cite{roemmele2011choice} proposes a benchmark for commonsense causal reasoning. It asks which of two alternatives has a causal relationship (either cause or effect) with a given premise. Finally, the temporal ordering of events is often studied along with typical times and duration \cite{6061471,granroth2016happens,liconstructing,zhou2019going}. 

\paragraph{Types of commonsense inferences.} While most commonsense work only pertains to non-situational semantic knowledge such as that captured by ConceptNet \cite{speer2017conceptnet}, in this paper we focus on commonsense based on naive psychology, a core human ability that allows people to reason about mental states such as reactions, intents, goals and beliefs \cite{heider1958} in particular situations. ATOMIC is specifically designed to capture such knowledge and we focus on such socially motivated commonsense, though our distant supervision approach and our proposed model are extensible to other knowledge bases and forms of commonsense. 


%% file: figures/task_examples.tex
{\setlength\tabcolsep{2pt} 
\begin{table*}[t]
\centering
\small
\begin{tabular}{llll}
\toprule
\textbf{Narrative}  & \textbf{Dimension} & \textbf{Discourse-Agnostic} & \textbf{Discourse-Aware}  \\ 
\midrule
Lenny was digging a hole in his yard to plant a tree. & \multirow{5}{*}{\texttt{PersonX} needed} & \multirow{5}{*}{to be in a pool \incorrect} & \multirow{5}{*}{to have a shovel \correct} \\
...\\ He jammed the shovel harder into the ground.\\ 
\textbf{All of a sudden water started spurting out of the hole.}\\  \midrule

Carla worked at the mall. & \multirow{5}{*}{\texttt{PersonX} needed} & \multirow{5}{*}{to be hungry \correct} & \multirow{5}{*}{to drive to the foodcourt \correct} \\ \textbf{For her lunch break she ate at the food court.} \\ ... \\ Carla's co-worker bought her lunch. \\ \midrule

Sports day was always Emma's favourite day at school. & \multirow{5}{*}{\texttt{PersonX} wants to} & \multirow{5}{*}{practice more \incorrect} & \multirow{5}{*}{to win \correct} \\ 
... \\ 
This year, a girl who moved to the school entered the 100m sprint.\\ 
\textbf{Emma had never seen her practice, so thought she would be fine.}\\ \midrule
The water bill at Ella's house had been high. & \multirow{5}{*}{\texttt{PersonX} wanted to} & \multirow{5}{*}{admire the view \incorrect} & \multirow{5}{*}{find a leak \correct} \\ 
... \\ 
\textbf{Ella walked around the property.}\\ 
\bottomrule
\end{tabular}
\caption{Examples generated from the models in this paper: a discourse-agnostic (sentence-level) baseline, vs. our discourse-aware \model{}. We highlight the sentence that each inference was generated for in \textbf{bold}. Inferences are marked as plausible (\correct) or implausible (\incorrect).}
\label{table:task_examples}
\end{table*}
}

%% file: figures/atomic_dimensions.tex
\begin{table}[t]
    \centering
    \begin{tabular}{lll} 
    \toprule 
    \textbf{Type} & \textbf{Dimension} & \textbf{Template} \\ 
    \midrule 
    \multirow{3}{*}{Causes} & xIntent & PersonX wanted $e_2$\\
    & xNeed & PersonX needed $e_2$\\
    & xAttr & PersonX is seen as $e_2$\\ 
    \midrule
    \multirow{6}{*}{Effects} & xWant & PersonX wants $e_2$\\
    & xEffect & PersonX is likely $e_2$\\ 
    & xReact & PersonX then feels $e_2$\\
    & oWant & PersonY wants $e_2$\\ 
    & oEffect & PersonY is likely $e_2$\\ 
    & oReact & Others then feel $e_2$\\
    \bottomrule 
    \end{tabular}
    \caption{Natural language templates for ATOMIC dimensions.}
    \label{table:atomic_dimensions}
\end{table}

%% file: 03-task.tex
Our work is motivated by the question: \textit{can NLP systems explicitly generate commonsense inferences, that a human might implicitly make while reading a narrative?} To tackle this question, we formalize and introduce the discourse-aware commonsense inference task.\footnote{We use the term \textit{discourse-aware} to refer to data/systems that use paragraph-level information. Similarly, \textit{discourse-agnostic} systems only use sentence-level information.}

Formally, given a narrative with $T$ sentences $\{S_1,S_2...S_T\}$, the goal is to generate a set of commonsense inferences for the nine inferential dimensions (Table \ref{table:atomic_dimensions}) for each sentence $S_i$. This set of inferences generated for $S_i$ must also be consistent with the entire narrative. 
Maintaining consistency with the full narrative context requires reasoning about the relationship between past and future events.

Table~\ref{table:task_examples} shows some examples of discourse-aware (paragraph-level) and  discourse-agnostic (sentence-level) inferences. Sentence-level inferences are often inconsistent with the narrative. For example, the inference that a character needed ``to be in a pool" when the earlier context shows they are gardening (first row in Table~\ref{table:task_examples}) or that a character wants to ``practice more" when it has been established they are confident in their own abilities (third row).

%% file: 04-data.tex
Sentence-level inferences (e.g. those obtained from \comet{}) are inadequate to train models for our proposed task and obtaining direct supervision of discourse-aware inferences may be prohibitively expensive or infeasible to collect in large quantities at an effective quality standard level.  
Therefore, we use distant supervision to loosely align sentences in a narrative
to their discourse-aware commonsense inferences. 
First, we obtain discourse-agnostic inferences from either the \comet{}~model or the ATOMIC knowledge base. Next, we filter out inferences that are inconsistent with the rest of the narrative (described in Section~\ref{sec:data_prune_events}). 
Thus, we obtain \textit{silver} standard training data for training models for our task. Additionally, we create a smaller-scale validation set by manually validating inferences through a crowdsourcing annotation task (Section~\ref{sec:supervision_eval}).

\subsection{Source of Narratives}
\label{sec:data_rs}

The basis for our dataset are English stories from the ROCStories corpus \cite{mostafazadeh-etal-2016-corpus}, which consists of 98K five-sentence stories authored by workers on Amazon Mechanical Turk. 
Understanding these stories requires commonsense and temporal inferences that we aim to capture. We split the original ROCStories train set into train, dev, and test sets in a 90/5/5 ratio.

\subsection{Discourse-agnostic Inferences}
\label{sec:data_atomic}

\begin{table*}[t]
\begin{center}
\resizebox{1\textwidth}{!}{%
\begin{tabular}{  l l c c c} 
\toprule
Narrative  & Inference & Relevant? & LM score \\
\midrule
\textbf{Natalie's favorite movie is The Wizard of Oz}... & PersonX wanted: to see the film & \correct & -3.384     \\
I was at the grocery store...\textbf{I see the lines were very long}... & PersonX then feels: relieved & \incorrect &  -3.408 \\ 
Jim wanted to learn Spanish. \textbf{He tried taking a class}...  & PersonY/Others want: to catch up & \incorrect & -3.602 \\
\textbf{Our building had a summer bbq party today.} The manager took photos... & PersonX wants: to enjoy the party & \correct & -3.915 \\
Chris realizes that he rarely watches cable TV anymore. \textbf{He calls}...\textbf{to cancel}... & PersonX wanted: to be a good customer & \incorrect & -3.952 \\
My grandparents lived in Alabama. We used to travel there...\textbf{I miss traveling there}... & PersonX is seen as: sad & \correct & -3.518 \\
\bottomrule
\end{tabular}
}
\caption{Examples from the distantly supervised dataset. We highlight the most relevant (i.e. potentially contradictory or supporting) sections in the story for each inference being considered. } 
\label{table:ws_examples}
\end{center}
\end{table*}

We aim to generate the types of commonsense inferences defined by the ATOMIC knowledge base \cite{Sap2019ATOMICAA}. We obtain discourse-agnostic inferences using either of the following approaches. 


\paragraph{Heuristic:} For each sentence $S_i$ in the story, we get an initial set of candidate inferences $R_i$ by extracting ATOMIC tuples, $<e_1, d, e_2>$, in which $e_1$ and $S_i$ share either noun phrases or verb phrases. We repurpose the ROUGE metric \cite{lin-2004-rouge} to measure the surface-level relevance of a particular event $e_1$ to a sentence $S_i$. 
Specifically, we compute the ROUGE-1 $F_1$ score, which considers unigrams, and keep the top 10 inferences with respect to the score for each sentence and dimension. 

\paragraph{Model-Based:} We use \comet{} to generate commonsense inferences for each sentence $S_i$ in the story. We use beam search with a beam size of 10 to obtain a set of inferences for each sentence and dimension combination. 

More details on the distant supervision data curation process are given in the Appendix. 
 
\subsection{From discourse-agnostic to discourse-aware inferences} 
\label{sec:data_prune_events}
The inferences obtained by both heuristic and model-based methods (Section~\ref{sec:data_atomic}) only consider one sentence at a time. 
To improve coherence with the rest of the narrative,  
we filter the inferences that have a low \textit{coherence} with the given narrative. Specifically, inspired by information theory \cite{Shannon1948TheMT,Hale2001APE}, we define coherence as a measure based on the cross entropy of the story tokens conditioned on a particular candidate knowledge inference. For a tuple $<e_1, d, e_2> \in R_i$ matched to a sentence $S_i$, and a language model $\Theta$, we compute the cross entropy loss of the tokens in the story, where $<d, e_2>$ follow $S_i$: $CE(S_1,...S_i,<d, e_2>,...S_5)$.\footnote{Here we define cross entropy loss as $CE(t_1,...,t_n) = -\frac{1}{n}\sum^n_{i=1} log_2 p_\Theta(t_i|t_1,...,t_{i-1}). $ } We use a transformer-based language model, and convert $<d, e_2>$ to natural language using hand-crafted templates shown in Table~\ref{table:atomic_dimensions}. 

In practice, we divide the dimensions into causes (\texttt{xNeed}, \texttt{xIntent}, \texttt{xAttr}) and effects (\texttt{xWant}, \texttt{xEffect}, \texttt{xReact}, \texttt{oWant}, \texttt{oEffect}, \texttt{oReact}). For cause inferences, we compute coherence with the previous and current sentences in the story. For effect inferences we use the full story. This allows us to effectively measure how well the extracted inferences may follow from past or predict future story events. 


To ensure an equal distribution of inferences across dimensions, we order inferences by coherence score and keep the top 5 inferences for each sentence and dimension type. This filtering step is designed to reduce the number of contradicting inferences in our distant supervision corpus.

\subsection{Validation Set}
\label{sec:supervision_eval}

We validate a subset of the development set through crowdsourcing to obtain a gold evaluation set. 
We used Amazon Mechanical Turk and asked workers to judge the relevance of inferences for a given sentence within a story, leaving the interpretation of relevance to the best judgement of annotators.\footnote{We restrict annotators to US only.} Generally, we found that annotators adhered to a strict definition of relevance in which ambiguous inferences that may still be relevant to the story context at some point in the story timeline are labeled as irrelevant. See Table~\ref{table:ws_examples} for examples. 

We randomly sampled 542 stories from the development set, and for each story we randomly selected a sentence and a dimension, and annotated the 5 inferences associated with them. We had 3 annotators judge each example, and used the majority vote to obtain a gold label. We filtered out low quality annotations by manually checking for workers with low inter-annotator agreement and frequently incorrect labeling.\footnote{These were done primarily by workers who spent less than 20 seconds on a HIT.} 

Our annotations yielded fair inter-annotator agreement of Fleiss' $\kappa=0.338$ \cite{fleiss1971measuring} (p-value $<$ .001). Despite the challenges of this task, this value is higher or comparable to prior work achieved for the evaluation of commonsense knowledge.\footnote{$\kappa=0.23$ in judging commonsense knowledge triplets in \cite{Feldman2019CommonsenseKM} and between $\kappa=0.289$ and $\kappa=0.483$ in commonsense story generation in \cite{Guan2020AKP}.} The final evaluation subset consists of 607 inferences, across all different dimensions, from 313 unique stories that were found to be relevant by multiple human annotators (34.29\% of the inferences judged).

%% file: 05-model.tex
\begin{figure*}[t]
\centering
\includegraphics[width=1\textwidth]{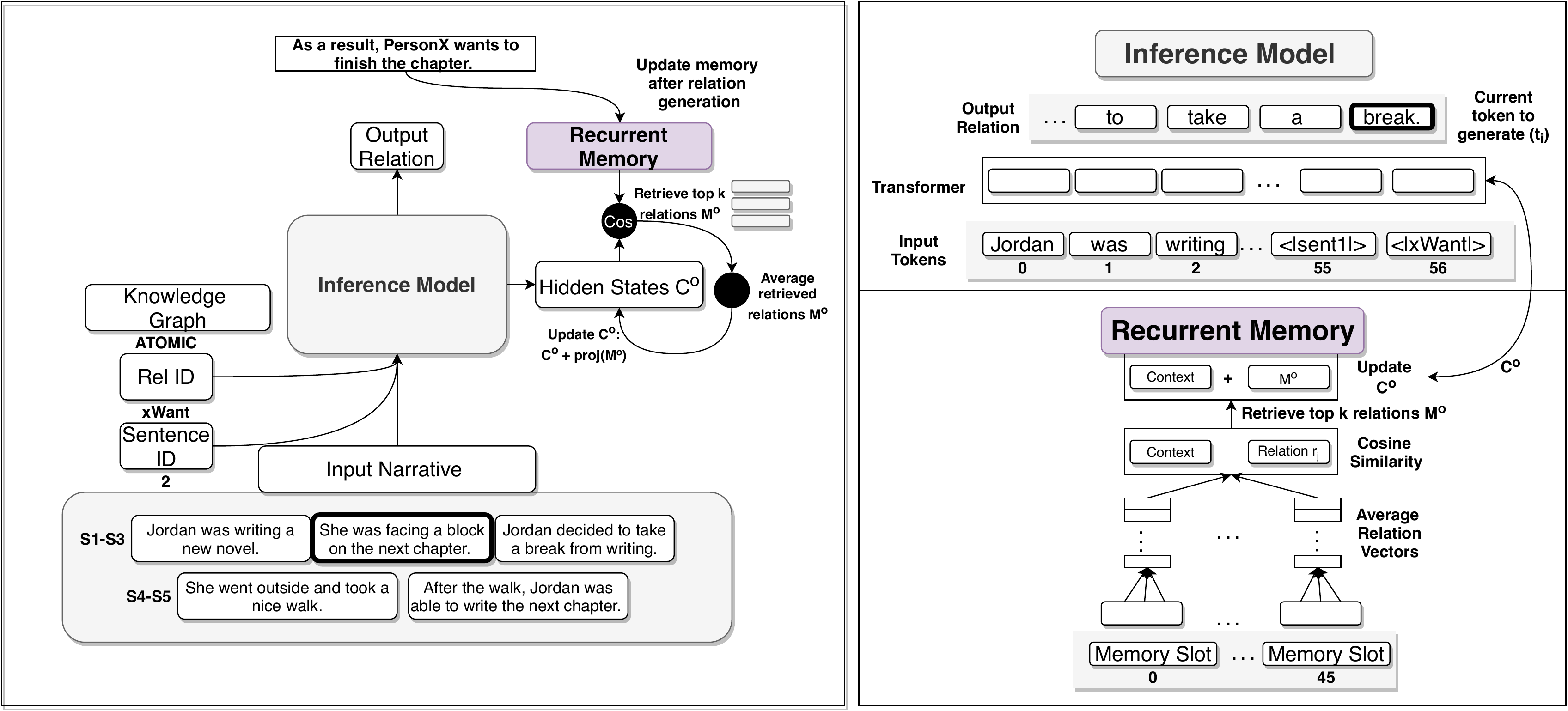}
\caption{An illustration of \model{} with a memory component. The model predicts an inference for a given sentence in the narrative (e.g., the second) and a requested ATOMIC dimension.}
\label{fig:model}
\end{figure*}

We draw inspiration from the distinction between semantic and episodic memory \cite{tulving}, and consider implicit commonsense knowledge in two ways: 1) \emph{semantic knowledge}, grounded in world knowledge and culture-specific social knowledge (e.g., ``leaks lead to high water bills''), and 2) \emph{episodic knowledge}, grounded in causal understanding and epistemic reasoning---i.e. reasoning that relates past events to current events (e.g., ``if a person gets a high water bill, they will want to find out why''). We introduce two variants of the \model{} controlled generation model: a memory-less model that focuses on semantic knowledge drawn from the context, and a model augmented with recurrent memory that allows us to explicitly incorporate episodic knowledge. 

Figure~\ref{fig:model} demonstrates generating inferences for a narrative using \model{} with recurrent memory. 


\paragraph{Memory-less Model.} 

Given a story context $c = \{S_1,S_2, \ldots ,S_T\}$ of $T$ sentences and a selected sentence $S_i$, we set the input to:
\begin{equation}
x = S_1~||~S_2~\ldots~S_T~||~s~||~d
\label{eq:input}
\end{equation}

\noindent where $s$ and $d$ are special tokens. $s$ represents the index of the selected sentence, while $d$ represents the required dimension in ATOMIC. $||$ denotes concatenation. In the example in Figure~\ref{fig:model}, the input provided to the model is:
\begin{equation*}
    x = \text{Jordan was writing...}~<|sent2|>~<|xWant|>
\end{equation*}


 We fine-tuned the base GPT and GPT2 transformer models \cite{Radford2019LanguageMA,Radford2018ImprovingLU} to generate the expected output, which is an inference for the dimension $d$ and sentence $S_i$. 

\paragraph{Memory-augmented Model.} To incorporate inferences generated for other sentences in the story while generating inferences for a given sentence, we extend the model with a recurrent memory component, inspired by episodic memory. $M^{m} \in \mathbb{R}^{R^m \times L^r \times H}$ is the external memory, where $R^m$ is either the maximum number of inferences per instance to store in memory (during training time) or the current number of instances (during decoding time), $L^r$ is the maximum inference sequence length,\footnote{We use a maximum memory size ($R^m$) of 45 inferences and a maximum sequence length of 100 tokens during training time. During decoding, we dynamically resize memory based on the number of inferences previously generated.} and $H$ is the hidden state dimension.

The memory-augmented model takes as input a memory update matrix $M^{u} \in \mathbb{R}^{R^u \times L^r \times H}$, where $R^u$ is the number of inferences used to update memory, and incorporates it into the memory matrix: 
\begin{equation}
    M^{m} = M^{m} \oplus f_{emb}(M^{u}) 
\label{eq:memory_update}
\end{equation}

\noindent $\oplus$ stands for matrix concatenation, and $f_{emb}$ is an embedding layer trained jointly with the rest of the model. After the memory is updated, we average $M^{m}$ across the token dimension to get $\theta^{mem} \in \mathbb{R}^{R^m \times H}$:
\begin{equation}
\theta^{mem} = \frac{1}{L^r} \cdot \sum_{l=1}^{L^r} {M^{m}}^l
\label{eq:avg_memory}
\end{equation}

We denote the context representation obtained from GPT or GPT2's hidden state as $C^{o} \in \mathbb{R}^{L^c \times H}$, where $L^c$ is the context sequence length. We average it across all tokens, obtaining $\theta^{ctx} \in \mathbb{R}^{H}$. We then prune the memory to the top-k most relevant inferences, measured by cosine similarity between the memory $\theta^{mem}$ and context vectors $\theta^{ctx}$. The memory output $M^{o} \in \mathbb{R}^ H$ is the average of the top-k inferences.

Finally, we reweigh the context representation $C^{o}$ to consider the memory:
\begin{equation}
     C^{o} = C^{o} + \operatorname{proj}(M^{o})
\label{eq:reweighing} 
\end{equation}

\noindent Where $\operatorname{proj}$ is a linear projection layer used to project the memory output into the same hidden dimensional space as the context representation.

At training time, the memory consists of previously extracted relations from our distant supervision, while at test time, it consists of previously generated inferences, recalling the model's prior decisions. For both \model{} model variants, we minimize the cross entropy loss of the entire sequence (input and output).

%% file: figures/human_eval_results.tex

\begin{table*}[t]
\centering
\small
\begin{tabular}{lllll}
\toprule
\textbf{Model} & \textbf{Decoding} & \textbf{True or Plausible (3-5) (\%)} & \textbf{Plausible (3)} (\%) & \textbf{Avg. Rating} \\ \midrule
\multirow{2}{*}{\comet{}} & greedy & 49.41 & 17.86 & 2.57 \\
& beam-10 & 63.69 & 26.19 & 2.93 \\
\midrule
PARA-H & beam-10 & 68.45 & 22.62 & 3.21 \\
PARA-H+mem & beam-10 & 66.67 & 27.98 & 3.05 \\
PARA-M & beam-10 & 74.40 & 23.81 & \textbf{3.44} \\
PARA-M+mem & beam-10 & \textbf{77.38} & \textbf{31.55} & 3.42 \\
\bottomrule
\end{tabular}
\caption{Human evaluation results. We highlight the overall best performing model in \textbf{bold}. All \model~results are using GPT2 models. } 
\label{table:human_eval_results}
\end{table*}


%% file: 06-experimental_setup.tex
\subsection{Training Setup}
\label{sec:training}

All models are implemented using the Transformers package \cite{Wolf2019HuggingFacesTS}, and trained for a maximum of 20 epochs. Training is performed using an Adam optimizer with linear warmup \cite{Kingma2015AdamAM}. We also simulate a batch size of 16 using gradient accumulation and an actual batch size of 4. For GPT2, the learning rate is $2*10^{-5}$ and for GPT we use a learning rate of $6.25*10^{-5}$. All other hyperparameters follow \cite{Radford2019LanguageMA,Radford2018ImprovingLU}. We retrieve the top $k=1$ inferences from memory.\footnote{For GPT2 we use memory during training and decoding. For GPT, we report results using training-only memory.} We use the 124M parameter version of the GPT2 model.

\subsection{Decoding Setup}
\label{sec:decoding}

For decoding, we use beam search with a beam size of $b \in \{1,10\}$. The maximum decoding length is 50 tokens. Unlike at training time, where we take a single dimension for each sentence in each story, at decoding time we generate inferences from every dimension for every sentence. For both training and decoding, all experiments are run using 64 Intel(R) Xeon(R) Gold 6130 x86-64 CPUs at 2.10GHz and a Quadro RTX 8000 GPU. 

\subsection{Baselines}
\label{sec:baseline}

As a baseline, we use the \comet{} model, pre-trained on ATOMIC, to generate sentence-level inferences for each sentence in the story.\footnote{See the original paper for details.} As an additional baseline, we use a retrieval model (BERT-KNN) based on the K-Nearest Neighbor search algorithm (k=1). We embed ATOMIC events using BERT \cite{Devlin2019BERTPO}, then find the closest ATOMIC event node for each story sentence to get a set of matching inferences.

%% file: 07-experiments.tex
We report the performance of all models for automatic evaluation and the top 6 model variations (two \comet{} variations and four \model{} variations) for human evaluation. For \model{}, we report the variants with and without memory, trained on either the heuristic matching approach (PARA-H) or the model-based approach (PARA-M), as described in Section~\ref{sec:data_atomic}. 

\subsection{Human Evaluation}
\label{sec:human_eval}

We follow a similar crowdsourcing setup to the validation presented in Section~\ref{sec:supervision_eval} to measure the quality of generated inferences. We sampled 336 inferences from 56 unique stories. We show crowdworkers the full story, a specified dimension, and a generated inference. We specify the assignment of \texttt{PersonX} to the syntactic subject of the sentence.\footnote{We manually corrected incorrect parses such as those in which the subject of the sentence is not a person.} 

Following \citet{Zhang2017OrdinalCI}, we ask workers to judge the likelihood of inferences based on a 5-point Likert scale: obviously true (5), generally true (4), plausible (3), neutral or unclear (2), and doesn't make sense (1). Table~\ref{table:human_eval_results} displays the percent of inferences judged as plausible or true (3-5), and plausible (3), and the average rating per inference (using majority voting). 

Overall, \model{} generations are scored with higher average ratings, between 3.05 and 3.44 points compared to 2.57 and 2.93 points for the \comet{} baseline variants. Specifically, the memory-augmented variants  produced notably more plausible inferences than any other model. We observed that inferences in this category tend to be less obvious---e.g. restating information from the context, producing generic inferences---and recover plausible implicit inferences. 



\subsection{Automatic Evaluation}
\label{sec:automatic_eval_metrics}

\input{figures/auto_eval}
\input{figures/blogposts}

\paragraph{Similarity to the gold inferences.} We follow the ATOMIC and \comet{} automatic evaluation setup using BLEU \cite{Papineni2001BleuAM}, which measures the n-gram overlap between the generated and gold inferences. 

\paragraph{Novelty.} Following \citet{Jastrzebski2018CommonsenseMA}, we compute novelty by measuring the percentage of generated inferences that do not appear verbatim in ATOMIC. We account for slight paraphrases by counting as novel the generated inferences that have an edit distance ratio of less than 0.95 with all ATOMIC events. 

\paragraph{Discourse-level Coherence.} We use natural language inference \cite[NLI;][]{dagan2013recognizing} as a proxy for measuring the narrative-level coherence of the predicted inferences. We define coherence as follows - at the very least, the story must not contradict any of the predictions, and it may possibly entail some of the predictions. We use the pretrained SemBERT model \cite{Zhang2019SemanticsawareBF}, a variant of BERT augmented with explicit semantic role labels, to compute NLI labels (\emph{entailment}, \emph{neutral}, \emph{contradiction}). 
\\
\\
Table~\ref{table:autoeval_gold} provides a summary of the automatic evaluation results on the gold subset. The \model{} variants outperform the sentence-level baselines across all metrics. The novelty results show that \model{} models are capable of generating inferences that did not appear in the original ATOMIC knowledge graph. The memory-augmented models generated inferences that were more coherent with the story, reducing the percents of contradicting inferences from 46.15\% (in \comet{}) to 40.58\%. We find the GPT models generally have comparable or better performance to GPT2 models on automatic metrics, which we hypothesize is due to the fact GPT was pretrained on story text and has specific knowledge pertaining to implicit knowledge underlying narratives. Overall, we find that incorporating narrative coherence through either episodic knowledge from the recurrent memory mechanism and/or context from other story events improves BLEU-1 by up to 3.33 points and BLEU-2 by up to 2.70 points.

%% file: figures/auto_eval.tex
\begin{table*}[t]
\centering
\small
\begin{tabular}{llrrrr} 
\toprule
 \textbf{Model} & \textbf{Decoding} & \textbf{BLEU-1} & \textbf{BLEU-2} & \textbf{Novelty} & \textbf{NLI} \\ 
 \midrule
\multirow{1}{*}{BERT-KNN} & - & 79.99 & 69.14 & - & 44.84\\
 \multirow{2}{*}{\comet{}} & greedy & 85.78 & 80.87 & 3.03 & 53.85 \\
 & beam-10 & 87.91 & 80.10 & 18.87 & 51.44
 \\ 
 \midrule
 \textsc{PARA}-H (GPT) & beam-10 & 91.00 & 83.14 & \underline{17.09} & 54.63 \\
 \textsc{PARA}-H+mem (GPT) & beam-10 & 90.99 & \underline{83.43} & 16.09 & 56.23 \\
 \textsc{PARA}-M (GPT) & beam-10 & 91.03 & 83.06 & 12.56 & 52.72 \\
 \textsc{PARA}-M+mem (GPT) & beam-10 & \underline{91.09} & 82.88 & 12.54 & \underline{\textbf{59.42}} 
 \\
 \midrule
 \textsc{PARA}-H ($\text{GPT}_2$) & beam-10 & 91.03 & 83.43 & 15.99 & 54.95 \\
 \textsc{PARA}-H+mem ($\text{GPT}_2$) & beam-10 & \textbf{\underline{91.24}} & \underline{\textbf{83.57}} & 14.39 & \underline{56.23} \\
 \textsc{PARA}-M ($\text{GPT}_2$) & beam-10 & 89.44 & 81.89 & \underline{\textbf{20.96}} & 54.63 \\
 \textsc{PARA}-M+mem ($\text{GPT}_2$) & beam-10 & 89.68 & 82.18 & 20.06 & 54.95 \\ \bottomrule
\end{tabular}
\caption{Performance according to the automatic evaluation metrics. The best performing model for a specific \model{} variant (GPT or GPT2) is \underline{underlined}. We highlight the overall best performing model in \textbf{bold}. The NLI score is the percent of stories for which the model predicted entail or neutral.}
\label{table:autoeval_gold}
\end{table*}


%% file: figures/blogposts.tex
\begin{table}[]
    \centering
      \resizebox{1\linewidth}{!}{  
    \begin{tabular}{l} \toprule
           \textbf{Story}: Almost a month ago now, the radio station got struck \\
           by lightning. It fried the router and the cable modem.\\
            \textbf{[Before $\text{\faBolt}$, PersonX wanted]}\\
           We got the new equipment right away. \\ \midrule
           \textbf{\comet{}} (beam-10): to be a good cook, to eat 
         \\ \midrule
         \textbf{PARA-M}: to have better internet, 
         to not be bothered \\ \bottomrule \toprule
         \textbf{Story}: I posted a moment ago regarding a girl I asked out...\\
           she said she would like to do something, \\
           but work made it difficult. That was a couple of weeks back... \\ 
           \textbf{[Next, PersonX will]}
           \\\midrule
           \textbf{\comet{}} (beam-10): get married,  get a divorce
         \\ \midrule
         \textbf{PARA-M}:  be asked out, get rejected \\ \bottomrule   
    \end{tabular}
    }
    \caption{Examples of personal blog posts with commonsense model predictions. Here we assume PersonX to be the narrator of the blog post.}
    \label{tab:examples_continued}
\end{table}


%% file: 08-case_study.tex


To test the ability of the model to generalize to more complex narratives requiring further pragmatic reasoning \cite{sap-etal-2020-social}, we sampled and manually evaluated a set of 111 story/sentence/dimension triplets from personal blog posts in the COSMOSQA machine reading comprehension test set \cite{Huang2019CosmosQM}. While these narratives tend to be of a similar or shorter length than ROCStories, they require more real-world understanding. They also contain nuanced descriptions of social interactions.

We found that our model is effective at predicting inferences in an unsupervised setting with 49.55 \% of relations labeled as true and 20.72\% of relations labeled as plausible (vs. 20.72\% and 27.03\% for \comet{}). We noticed that our model more frequently overcomes two major plausibility errors in unsupervised commonsense inference - \textit{off-topic} and \textit{temporarily inappropriate} predictions (see Table \ref{tab:examples_continued}). 
For example, our model is able to correctly predict the likely intentions of someone owning a router and cable modem (example 1), while \comet{} predictions incorrectly focus on meal preparation. \comet{} also sometimes makes relevant but farfetched predictions while our model's inferences are better situated within a narrative timeline (example 2). 





%% file: 10-conclusion.tex

We introduced a new task of discourse-aware commonsense inference over narratives. To target this task, we proposed a new model, \papername{}, trained using distant supervision, that captures narrative discourse.

Despite the challenges of the task, we demonstrated the effectiveness of our approach using both automatic and human evaluations. In particular, our models were able to generate more implicit and novel discourse-aware inferences. In the future, we are interested in exploring further extensions of our work to downstream paragraph- and narrative-level tasks that may benefit from access to commonsense knowledge.

%% file: 11-ethics.tex
We note that the knowledge represented by current resources captures a mix of \textit{general} commonsense which the majority of readers would find likely regardless of background (including factual commonsense) and \textit{culturally-specific} commonsense which is only likely to some readers (i.e. is not the most logical conclusion for all readers). 
%
One likely contributor to this specificity of commonsense is the dependency on online crowdsourcing for annotation and generation of commonsense knowledge. A 2016 report from Pew Research\footnote{https://www.pewresearch.org/internet/2016/07/11/turkers-in-this-canvassing-young-well-educated-and-frequent-users/} found that a sample of MTurk crowd-sourcing workers was heavily skewed demographically. This has an unintended side effect of enforcing a potentially harmful assumption that the ``commonsense knowledge" is only knowledge agreed upon by a specific demographic or cultural majority. 
 Proposed steps for future work on Discourse-Aware Commonsense Inference include:
\begin{itemize}
    \item Multicultural and multilingual commonsense datasets that capture a more distributional view of commonsense, allowing for both sociocultural overlap and disagreement about likelihood of relevant inferences. 
    \item New evaluation metrics and frameworks for commonsense that consider likelihood rather than hard labels of relevancy. We begin to explore this with our categorical labeling of extracted commonsense knowledge. 
    \item Social commonsense inference models that consider multiple audience points-of-view. 
\end{itemize}

%% file: 12-appendix.tex
\appendix
\section{Appendices}
\label{sec:appendix}

\subsection{Efficient Collection of Distant Supervision}
\label{sec:appendix_distant_supervision}

To make the collection of distant supervision more efficient, we parallelize the processing of candidate inferences, computing the language model scores in batches of 130 candidate inferences. We split the process across 3 GPUs, with 30000 instances handled by each machine. We also cache the extracted verb and noun phrases. The average runtime for 30,000 instances is approximately 5 hours for the heuristic distant supervision and 50 hours for the model-based distant supervision. 

\subsection{Details of Evaluation Set} 

\begin{figure}
    \centering
    \includegraphics[width=1\linewidth]{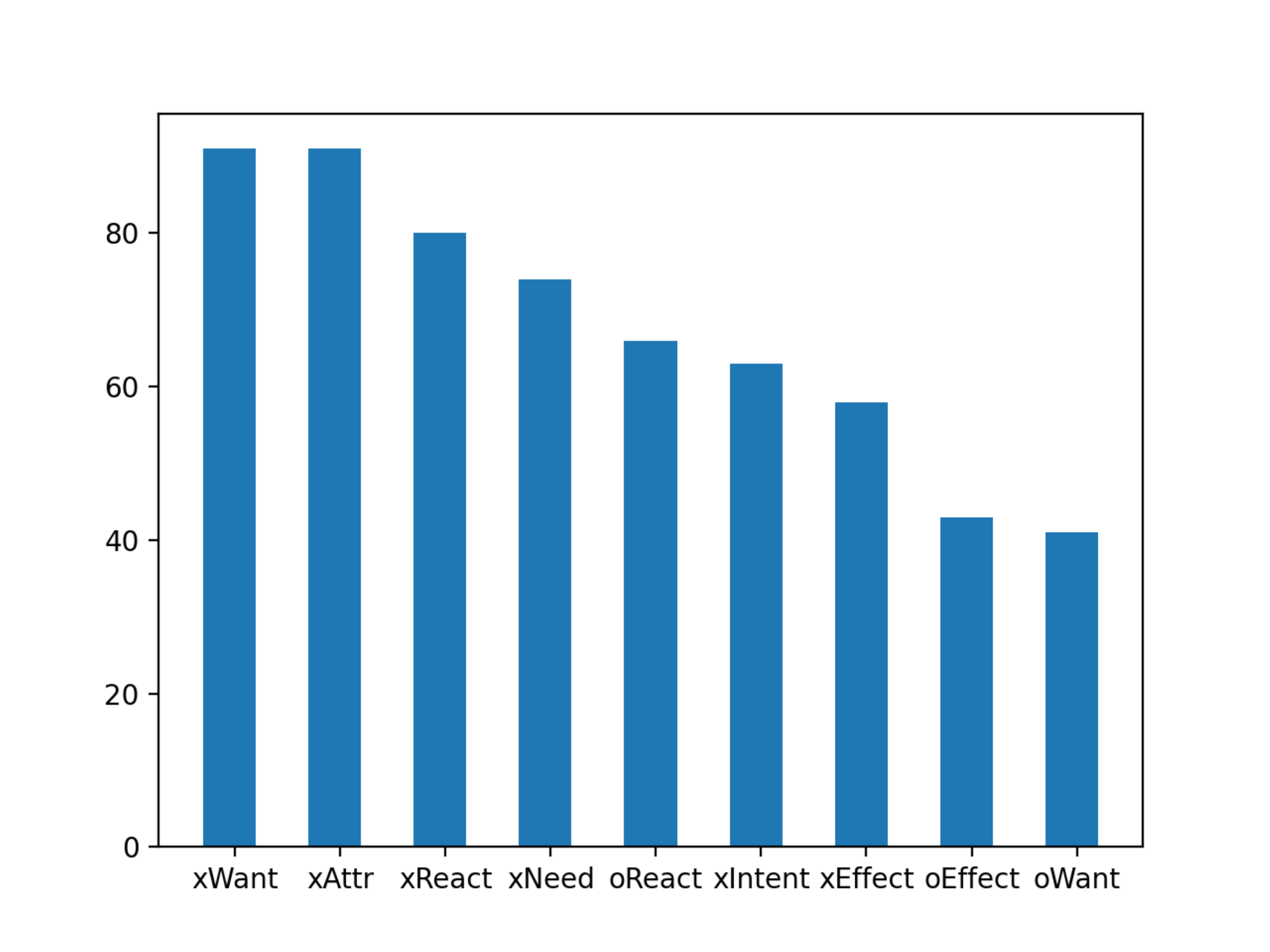}
    \caption{The distribution of ATOMIC relation dimensions in human-validated data.}
    \label{fig:eval_distr}
\end{figure}

We provide the distribution of ATOMIC knowledge graph dimensions covered by the validation set in figure \ref{fig:eval_distr}. During evaluation, we use the full beam of inferences generated for a story/sentence/dimension triplet. For BLEU and novelty we consider the full beam. For NLI we consider the top relation. BLEU scores are computed from inferences with templates, while the relative ordering of model performance is unchanged by removing templates this does lower individual BLEU scores. Since the SemBERT NLI model was trained on sentence-level NLI tasks, we use each sentence of a narrative as the premise and a generated inference as the hypothesis, then aggregate across sentences. We also normalize the inferences to more closely match the language the model was trained on, using hand-crafted templates.\footnote{We removed colons from the templates in Table~\ref{table:atomic_dimensions}, lowercased the inferences, converted verbs to their infinitive form using WordNet, and replaced \texttt{PersonX} with the pronoun ``they''. }

\subsection{Additional Training Details}
\label{sec:appendix_training}

We added the tokens shown in Table \ref{table:special_tokens} to the base GPT2 model. For GPT, we additionally added an `$<|$endoftext$|>$' eos token. 

\input{figures/special_tokens}

The base \model{} GPT2 models are trained for 8 epochs, which took approximately 1.5 hours per epoch. We train the GPT base models for 2 epochs. The memory-augmented GPT2 models are trained for 9 epochs, with approximately 2 hours per epoch. The GPT2 models are trained for 3 epochs. We used the pre-trained \comet{} model with 116M parameters. 

We manually tuned the GPT2 learning rate based on the validation loss, testing values in the range of [.2,.00002]. All other hyperparameters follow the specifications in the original GPT2 paper. For GPT, we use a learning rate of 6.25e-05. All results are from one trained version of each model and run with a fixed random seed.  

GPT2 was pretrained on WebText. The GPT model was pretrained on the BooksCorpus. 

\subsection{Additional Memory Augmentation Details}
\label{sec:memory_details}

Since memory is initially empty, we only use memory-augmented decoding at inference time after at least 1 inference has been generated for a particular story/sentence pair. We found that using memory augmentation at both training and decoding time improves results across metrics for GPT2, but only improves NLI and novelty of generated inferences for GPT while decreasing BLEU. All reported results for GPT only using memory during training time. 

\subsection{Additional Human Evaluation and Annotation Details}
\label{sec:appendix_annotation}

We paid the annotators \$0.25 per MTurk HIT, which we estimate is equivalent to a payrate of \$10-\$15 per hour. We encouraged annotators to leave feedback and comments. To ensure the quality of annotations, we required that annotators be located in the US and have at least 98\% approval rate on 1,000 prior HITs.

\begin{figure*}
    \centering
    \includegraphics[width=1\linewidth]{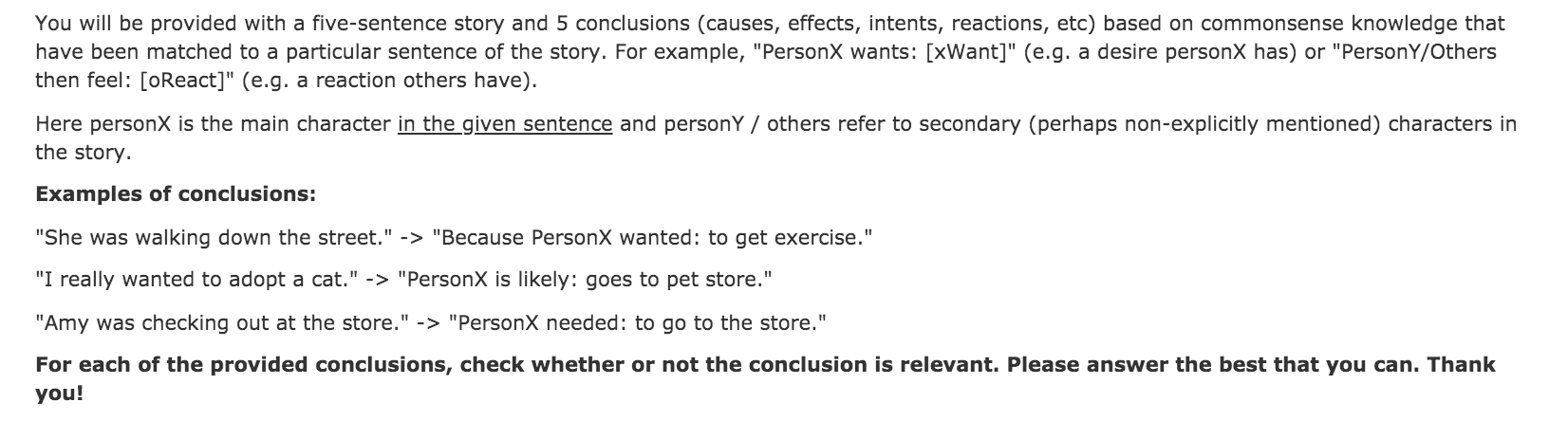}
    \caption{Annotation instructions for the relevancy task.}
    \label{fig:relevance}
\end{figure*}

In the validation task described in Section~\ref{sec:supervision_eval}, we instructed the workers to judged the relevance of the inferences as detailed in Figure~\ref{fig:relevance}. 

In the human evaluation task described in Section~\ref{sec:human_eval}, we presented the workers with the following descriptions of the Likert scale, exemplified with the sentence ``I really wanted a dog.'' (we used the term ``conclusion" rather than ``inference'' to simplify the terminology):

\begin{enumerate}
    \item \textbf{Doesn't Make Sense}: The conclusion clearly contradicts the story or is nonsensical given the context.\\
    \underline{Example:} ``PersonX wanted: colorless green ideas.'' 
    \item \textbf{Unclear or Neutral}: The conclusion is completely unrelated to the story context, though not necessarily contradictory. Additional information is needed.\\
    \underline{Example:} ``PersonX wanted: a cat.''
    \item \textbf{Plausible}: The conclusion could be relevant, but is not directly related or slightly off-topic. \\
    \underline{Example:} ``PersonX needed: to move to a new house.'' (Explanation: pets require more space and it's possible PersonX's house is too small for pets).
    \item \textbf{Generally True}: The conclusion is not stating information already in the story context, but it follows from the story and is very probable.\\ 
    \underline{Example:} ``PersonX needed: to go to the pet store''.
    \item \textbf{Obviously True}: The conclusion is clearly relevant to the story context. These conclusions restate information from the original story context.\\
    \underline{Example:} ``PersonX wanted: a dog''. 
\end{enumerate}

\input{figures/full_human_eval_results}

Table~\ref{table:full_human_eval_results} displays the full human evaluation results and Table \ref{table:gen_examples} shows additional generated examples. 

\input{figures/gen_examples}

\subsection{Case Study Details} 

To adapt the dataset to our task, we keep the narrative contexts but leave out aligned question/answer pairs. We restricted the set to five-sentence narratives, truncating longer narratives, to use our models pretrained on the ROCStories distant supervision data.

%% file: figures/special_tokens.tex
\begin{table}[t]
    \centering
    \begin{tabular}{ll} 
    \toprule 
    \textbf{Type} & \textbf{Token}  \\ 
    \midrule 
    \multirow{5}{*}{Sentence-Level Special Tokens} & \texttt{<|sent0|>} \\
     & \texttt{<|sent1|>}\\
    & \texttt{<|sent2|>}\\ 
     & \texttt{<|sent3|>}\\
     & \texttt{<|sent4|>}\\
    \midrule
    \multirow{9}{*}{ATOMIC Special Tokens} & \texttt{<|xNeed|>}\\
    & \texttt{<|xIntent|>} \\
    & \texttt{<|xWant|>} \\
    & \texttt{<|oEffect|>}\\
    & \texttt{<|xReact|>}\\
    & \texttt{<|oWant|>}\\
    & \texttt{<|oReact|>}\\
    & \texttt{<|xEffect|>}\\
    & \texttt{<|xAttr|>}\\ 
    \midrule
    \multirow{1}{*}{Additional Special Tokens} & \texttt{<|PAD|>}\\
    \bottomrule 
    \end{tabular}
    \caption{Special tokens used.}
    \label{table:special_tokens}
\end{table}

%% file: figures/full_human_eval_results.tex
\begin{table*}[t]
\centering
\small
\begin{tabular}{llrrrrrr}
\toprule
\textbf{Model} & \textbf{Decoding} & \textbf{Obviously} & \textbf{Generally} & \textbf{Plausible (\%)} & \textbf{Neutral (\%)} & \textbf{Doesn't Make} & \textbf{Avg. Rating} \\
& & \textbf{True (\%)} & \textbf{True (\%)} & & & \textbf{Sense (\%)} \\
\midrule
\multirow{2}{*}{\comet{}} & greedy & 16.07 & 15.48 & 17.86 & 10.71 & 39.88 & 2.57 \\
& beam-10 & 17.26 & 20.24 & 26.19 & 10.71 & 25.60 & 2.93 \\
\midrule
PARA-H & beam-10 & 22.02 & 23.81 & 22.62 & 16.07 & 15.48 & 3.21 \\
PARA-H+mem & beam-10 & 15.48 & 23.21 & 27.98 & 17.26 & 16.07 & 3.05 \\
PARA-M & beam-10 & 30.95 & 19.64 & 23.81 & 13.69 & 11.91 & 3.44 \\
PARA-M+mem & beam-10 & 28.57 & 17.26 & 31.55 & 13.10 & 9.52 & 3.42 \\
\bottomrule
\end{tabular}
\caption{Full human evaluation results.} 
\label{table:full_human_eval_results}
\end{table*}

%% file: figures/gen_examples.tex
\begin{table*}
\centering
\resizebox{\textwidth}{!}{%
    \begin{tabular}{ l  c   c  c }
    \toprule
    Story  & Dimension & Discourse-Agnostic & Discourse-Aware  \\ \midrule
 Sports day was always Emma's favourite day at school. \\ She was the fastest 100m runner for 3 years in a row. \\ This year, a girl who moved to the school entered the 100m sprint.\\ \textbf{Emma had never seen her practice, so thought she would be fine.}\\ The girl beat Emma by over a second. & PersonX wants to: & practice more \incorrect & to win \correct \\ \midrule
The boy opened the cereal box. \\ He heard a noise in it. \\ \textbf{He reached inside and felt around.} \\ He found a toy. \\ The boy played with that toy all day. &  PersonX wanted to: & find something \correct & to find something \correct  \\ \midrule
The man gave a presentation. \\ He asked for questions from the audience. \\ \textbf{The audience had a lot of  questions.} \\ The organizer realized the man was a good presenter. \\ He asked him to present at the next convention. & PersonX is likely: & to get yelled at \incorrect & learn something \correct \\ \midrule
Ed was performing in a band. \\ They left on their first tour. \\ \textbf{He was nervous about the travel}. \\ But the band loved being on the road. \\ They extended their tour for many more months!& PersonX needed: & to book the tickets \correct & to go to the airport \correct \\ \midrule 
Carla worked at the mall. \\ \textbf{For her lunch break she ate at the food court.} \\ One day she forgot her money. \\ Her co-worker found out she couldn't eat. \\ Carla's co-worker bought her lunch. & PersonX needed: & to be hungry \correct & to drive to the foodcourt \correct
\\ \midrule
Braden is a teacher who has been let go due to downsizing. \\
\textbf{Braden went home to tell his wife and to start job searching.} \\
After months of no interviews Braden decided to blog about it. \\
Turns out people enjoyed reading about his rants online. \\
Braden began a professional blog which is now his decent paying job. & PersonX needed: & to call his wife \correct & to find a job \correct  \\ \midrule
Buddy was a dare devil. \\ \textbf{His friends told him if he didn't stop he'd be sorry.} \\ Buddy ignored his friends and kept on doing dangerous stuff. \\ One day buddy tried to jump off the top of a house. \\ He landed on the ground instead of the trampoline and broke his legs. & PersonX seen as: & remorseful \incorrect & caring \correct \\ \midrule 
Tom walked into his classroom one morning and saw a new tank. \\ Inside of the tank, there was a little turtle. \\ \textbf{Tom's teacher explained that the turtle was their new pet.} \\ She told the students that they could name the new turtle. \\ The students chose the name Henry. & PersonX is likely to : & get bitten by the turtle \incorrect & learn something new \correct  \\ \midrule
Lenny was digging a hole in his yard to plant a tree.\\ He hit something hard.\\ He jammed the shovel harder into the ground.\\ \textbf{All of a sudden water started spurting out of the hole.}\\ He had broken through a water pipe. & PersonX needed: & to be in a pool \incorrect & to have a shovel \correct \\ 
    \bottomrule
    \end{tabular}
}
\caption{Examples generated from the models in this paper: a discourse-agnostic sentence-level baseline, vs. our discourse-aware \model{}. We highlight the sentence that each inference was generated for in \textbf{bold}. Inferences are marked as plausible (\correct) or implausible (\incorrect). We note that in example \#6 and some other xNeed examples the xNeed dimension refers to an action or condition that is both prior and ongoing (e.g. ``Braden needed/still needs to find a job") rather than a prior action or condition that personX completed.}
\label{table:gen_examples}
\end{table*}